\documentclass[conference]{IEEEtran}
\IEEEoverridecommandlockouts
\usepackage{cite}
\usepackage{amsmath,amssymb,amsfonts}
\usepackage{algorithmic}
\usepackage{graphicx}
\usepackage{textcomp}
\usepackage{xcolor}
\usepackage{hyperref}
\usepackage{caption}
\usepackage{subcaption}

\def\BibTeX{{\rm B\kern-.05em{\sc i\kern-.025em b}\kern-.08em
    T\kern-.1667em\lower.7ex\hbox{E}\kern-.125emX}}

\begin{document}

\title{Docking-based Virtual Screening with Multi-Task Learning}

\makeatletter
\newcommand{\linebreakand}{%
  \end{@IEEEauthorhalign}
  \hfill\mbox{}\par
  \mbox{}\hfill\begin{@IEEEauthorhalign}
}
\makeatother

\author{\IEEEauthorblockN{Zijing Liu}
\IEEEauthorblockA{\textit{Baidu Inc.} \\
Shenzhen, China \\
liuzijing01@baidu.com}
\and
\IEEEauthorblockN{Xianbin Ye}
\IEEEauthorblockA{\textit{Jinan University} \\
Guangzhou, China \\
yexianbin@stu2019.jnu.edu.cn}
\and
\IEEEauthorblockN{Xiaomin Fang}
\IEEEauthorblockA{\textit{Baidu Inc.} \\
Shenzhen, China \\
fangxiaomin01@baidu.com}
\and 
\IEEEauthorblockN{Fan Wang}
\IEEEauthorblockA{\textit{Baidu Inc.} \\
Shenzhen, China \\
wangfan04@baidu.com}
\linebreakand
\IEEEauthorblockN{Hua Wu}
\IEEEauthorblockA{\textit{Baidu Inc.} \\
Beijing, China \\
wu\_hua@baidu.com}
\and
\IEEEauthorblockN{Haifeng Wang}
\IEEEauthorblockA{\textit{Baidu Inc.} \\
Beijing, China \\
wanghaifeng@baidu.com}
}

\maketitle

\begin{abstract}
Machine learning shows great potential in virtual screening for drug discovery. Current efforts on accelerating docking-based virtual screening do not consider using existing data of other previously developed targets. 
To make use of the knowledge of the other targets and take advantage of the existing data, in this work, we apply multi-task learning to the problem of docking-based virtual screening. With two large docking datasets, the results of extensive experiments show that multi-task learning can achieve better performances on docking score prediction. By learning knowledge across multiple targets, the model trained by multi-task learning shows a better ability to adapt to a new target.
Additional empirical study shows that other problems in drug discovery, such as the experimental drug-target affinity prediction, may also benefit from multi-task learning. 
Our results demonstrate that multi-task learning is a promising machine learning approach for docking-based virtual screening and accelerating the process of drug discovery.
\end{abstract}



\section{Introduction}
Bringing a new drug to market is time-consuming and expensive. The pipeline of drug discovery and development is long and complex, with a high failure rate. The cost of a new drug was estimated to be \$2.8 billion, and it took more than ten years~\cite{dimasi2016innovation}.
Enormous efforts have been made to accelerate this process and lower the cost-to-market. Computer-aided drug discovery (CADD) has long been one of the most promising directions. 
One initial step of drug discovery is to identify new drug-like chemicals which has a strong affinity to a predefined protein target (\autoref{fig1ab}~(a)). However, due to the high cost and time limit, it is impossible to perform chemical experiments to screen the massive chemical space. In CADD, computers are used to perform a preliminary virtual screening of a large number of compounds and select the most promising compounds for further chemical experiments~\cite{shoichet2004virtual}.
A common approach for virtual screening is to use molecular docking to predict the binding affinity and bound conformation. It has been shown that structure-based docking can achieve high hit-rates on ultra-large chemical libraries of more than 100 million compounds~\cite{lyu2019ultra,stein2020virtual}.

However, the commercially available compound libraries have been growing, and their sizes have become so large that the computational time of virtual screening emerges to be an issue. For example, ZINC, a widely used compound library, now has more than 1.3 billion purchasable compounds containing 736 million lead-like molecules~\cite{irwin2020zinc20}. With a platform for ultra-large virtual screening~\cite{gorgulla2020open}, docking one billion compounds needs more than 4 million CPU-hours or 173 days with 1000 CPUs (15 seconds/ligand). The high requirement of such computational resources brings difficulties for many research and drug discovery organizations. 

In order to mitigate the computational cost of growing virtual libraries, many new computational methods have been developed based on machine learning. A common approach is that a machine learning model is trained as a surrogate model to predict the output of molecular docking using the docking data from a small set of the compound library (\autoref{fig1ab}~(b)). The machine learning model is then used as a preliminary filter, and the top compounds predicted by the model are chosen for further validation. 

\begin{figure*}[h!]
\centering
 
\includegraphics[width=0.95\textwidth]{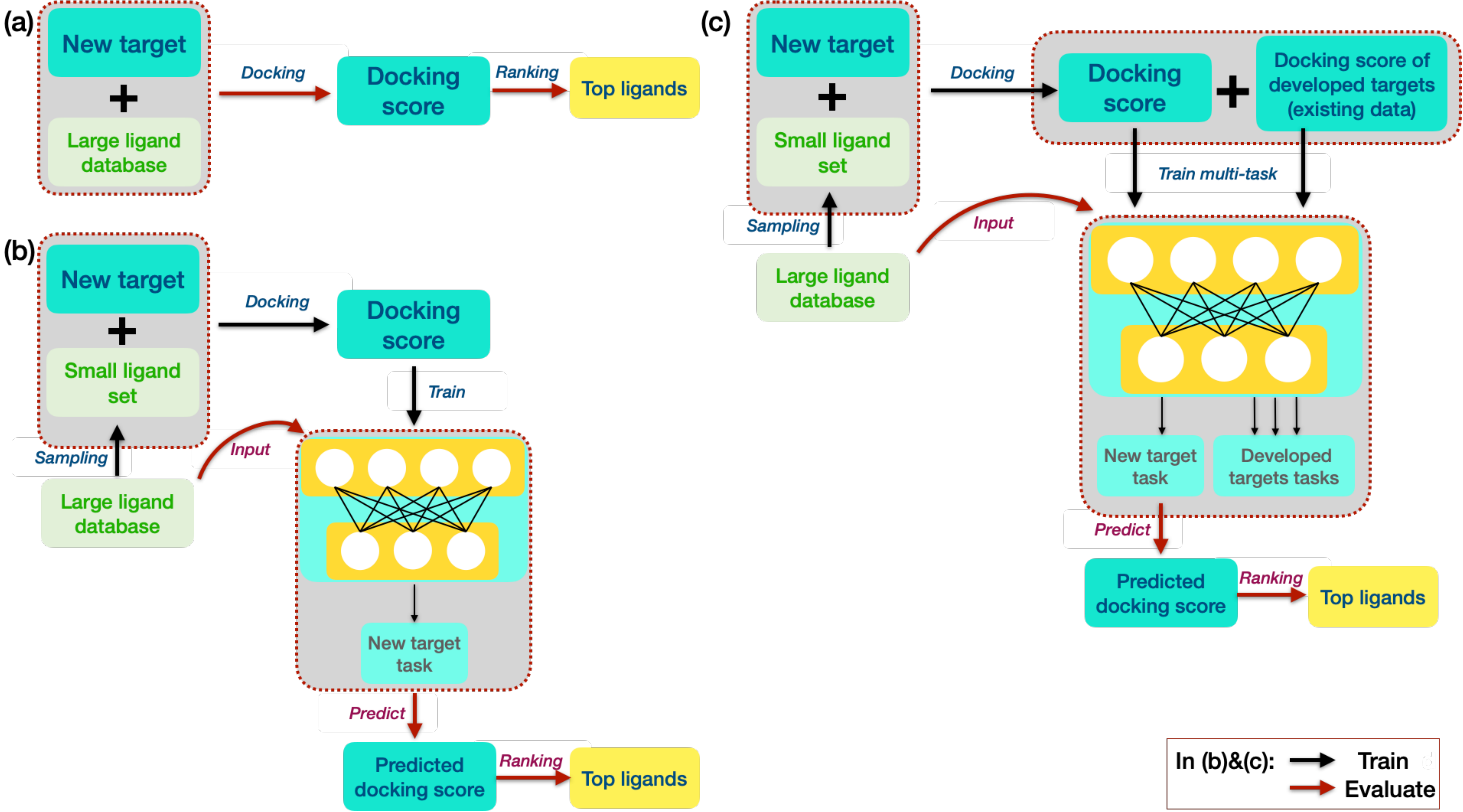}

\caption{The processes of docking-based virtual screening of a new target using (a) direct exhaustive docking, (b) machine learning accelerated screening, which samples a small subset of the ligands and use docking to obtain the data, and then train a machine learning model to predict the docking score of all the compounds in the large ligand database, (c) multi-task learning for screening, where the existing data of the other targets are combined with the new target docking scores to jointly train the model. Note that for active learning, the sampling and training processes in (b) may iterate for multiple rounds, but it does not use any other existing data.}
\label{fig1ab}
\end{figure*}

To further improve the accuracy of the machine learning model, various machine learning techniques have been applied to the problem of docking-based virtual screening. One approach that attracts a lot of interest in drug discovery~\cite{reker2019practical} is active learning, with applications in docking~\cite{gentile2020deep,graff2021accelerating,yang2021efficient} and free energy estimation~\cite{smith2018less, konze2019reaction}. 
Although active learning has been shown able to reduce the computational cost to screen a large library, it suffers from the bias introduced by the selection process~\cite{farquhar2021on}. 
More importantly, current machine learning methods for docking-based virtual screening, including recent works on active learning ~\cite{gentile2020deep,graff2021accelerating,yang2021efficient}, try to train the model with only the data of the new target, neglecting any existing data of other targets (Figure~\ref{fig1ab}~(b)).
In practice, a research organization or a pharmaceutical company, which is going to perform virtual screening on a new target, usually has a lot of related data of other targets accumulated from its past drug development projects.
In such a scenario, it is essential to take full advantage of the available data and avoid wasting data resources. As more and more data will be produced over time, these related data will play an increasingly important role.


The structure and property of a compound are closely related, which can be shown by the various applications of the quantitative structure-activity relationship (QSAR) or quantitative structure-property relationships (QSPR) models in drug discovery~\cite{nantasenamat2010advances}.
Docking is essentially a structure-based CADD approach.
The docking score prediction models learned from other targets may contain useful knowledge about the new target. From this insight, it is reasonable to jointly model the docking data of the new and previous targets, which leads us to use Multi-Task Learning (MTL)~\cite{zhang2021survey}.
MTL aims to learn multiple related tasks jointly such that a task can use the knowledge contained in other tasks to improve its performance, such as predictive accuracy and generalization ability. In previous studies, MTL has been mostly used for QSAR studies~\cite{xu2017demystifying,ramsundar2017multitask,lee2019silico,sosnin2019survey}.
In this paper, we investigate MTL in docking score prediction for the purpose of virtual screening (Figure~\ref{fig1ab}~(c)). 
We show that MTL improves the accuracy of docking score prediction.
By learning the docking scores of the new and other (developed) targets together, the trained model has a large boost and outperforms both single-task learning and active learning.
Our study also reports that the model trained with MTL learns a better representation of the compounds by sharing knowledge across multiple tasks, and thus can be used to adapt to a new task and achieve better performances.
Besides docking score prediction, we also apply MTL to drug-target affinity prediction, showing that MTL can be employed in other problems of drug discovery.

The contributions of our paper can be concluded as follows: 
\begin{itemize}
    \item By integrating the existing docking data, we use multi-task learning in the process of docking-based virtual screening.
    \item Empirical studies show that multi-task learning can have better performances than single-task machine learning and active learning with the same number of docking compounds. 
    \item With multi-task learning, the model can learn the common knowledge from related tasks and adapt to a new task with better performances.
    \item Multi-task learning is also applicable to other low data tasks in drug discovery, such as drug-target affinity prediction.
\end{itemize}

\section{Methods}
In this section, we introduce the MTL for virtual screening, where the targets with existing docking data are used to jointly train the model. The compound representation and neural network model are also described.

\subsection{Multi-task learning}
Given $m$ tasks $\{ \mathcal{T}_1,\mathcal{T}_2,\ldots,\mathcal{T}_m \}$, the aim of MTL is to learn all the tasks together to improve the performance for each task via learning the knowledge in other tasks. In the case of docking score prediction, one significant difference from traditional MTL is that we aim to improve the accuracy of predicting a new protein target, which we denote by $\mathcal{T}_0$, using the information learned from data of $m$ existing targets. Those tasks from existing targets are denoted as $\{ \mathcal{T}_1,\mathcal{T}_2,\ldots,\mathcal{T}_m \}$.

Nowadays, most of the models are based on deep learning due to their representation power and predictive ability. With deep neural networks, sharing the knowledge from other tasks can be easily achieved by parameter sharing. In a neural network model, some layers have shared weights across the tasks to jointly learn the knowledge from all the tasks, and there are also layers with the task-specific weights to capture the task-dependent features. The neural network model (\autoref{fig:mtl1}) can be represented by continuous and differentiable parametric functions with input $x$:
\begin{align}
    & z = G_{\phi}(x), \\
    & \Tilde{y}_i = F_{\theta_i}(z) \,,\, i=0,1,\ldots,m,
\end{align}
where $G_{\phi}$ is the layers with shared parameters $\phi$, $F_{\theta_i}$ is the layers with task-specific weights $\theta_i$, and $\Tilde{y}_i$ is the predicted output for task $i$.
The parameters $\phi$ and $\theta_i$ can be learned by optimizing a loss function $L(\Tilde{y}_i, y_i)$ where $y_i$ is the true label of task $i$. In virtual screening or docking score prediction, it is a regression problem, and the mean-squared error (MSE) loss is used. Note that a compound may have no label for some tasks. Let $\mathcal{X}_i$ denote the set of compounds that have the label of task $i$.
The total loss can be written as
\begin{align}
    \sum_{i=0}^m \sum_{x\in \mathcal{X}_i} L(\Tilde{y}_i, y_i),
\end{align}
which is minimized to learn the parameters of the neural networks.

\begin{figure}[!ht]
\centering
\includegraphics[width=0.55\linewidth]{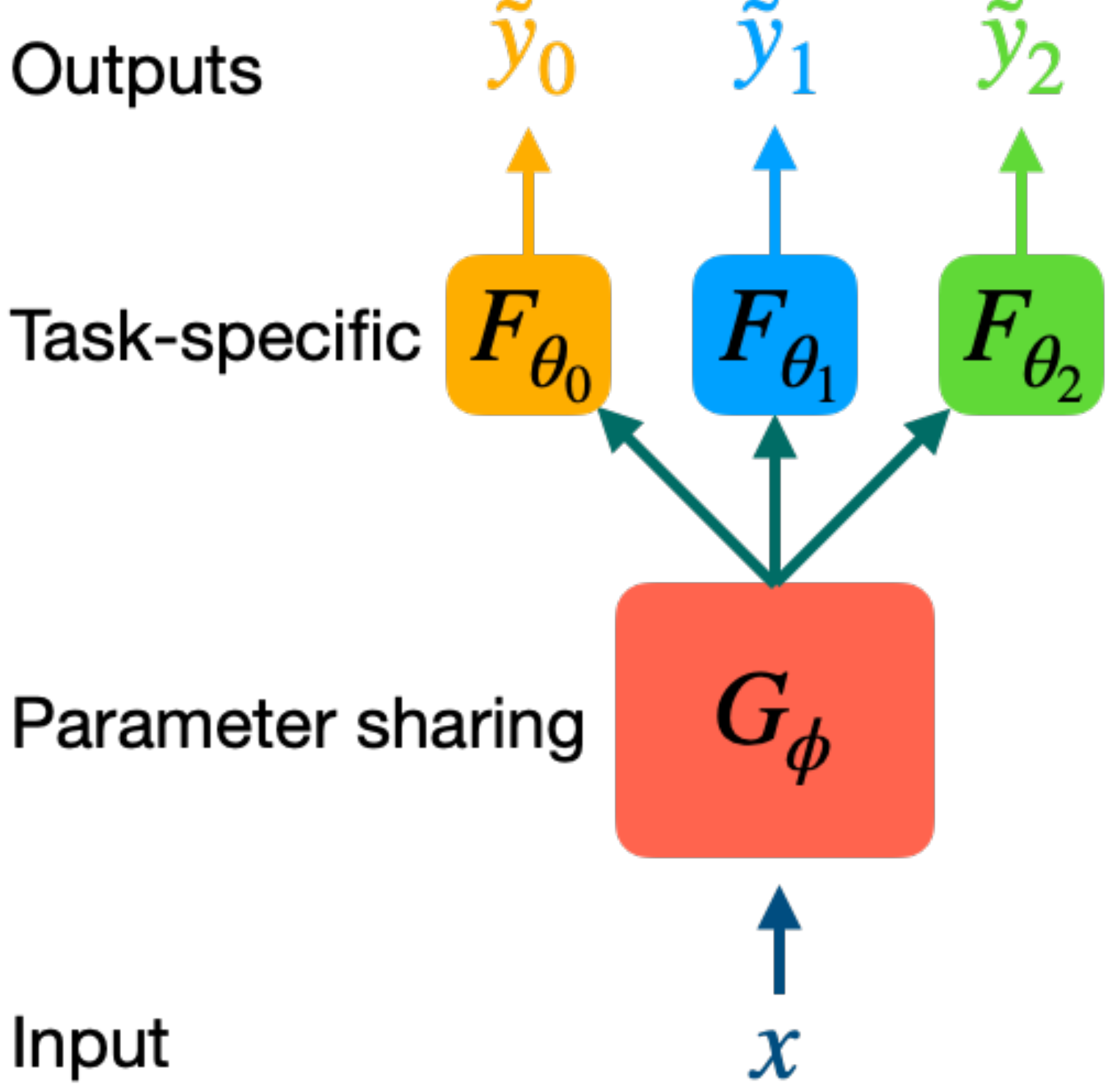}
\caption{MTL with shared-parameter and task-specific-parameter neural network layers. $\phi$ and $\theta_i$ are learnable parameters.}
\label{fig:mtl1}
\end{figure}

\subsection{MTL model for docking score prediction}
We now introduce the MTL model for docking score prediction in detail (Figure~\ref{fig:mtl2}).



\begin{figure}[!ht]
\centering
\includegraphics[width=0.65\linewidth]{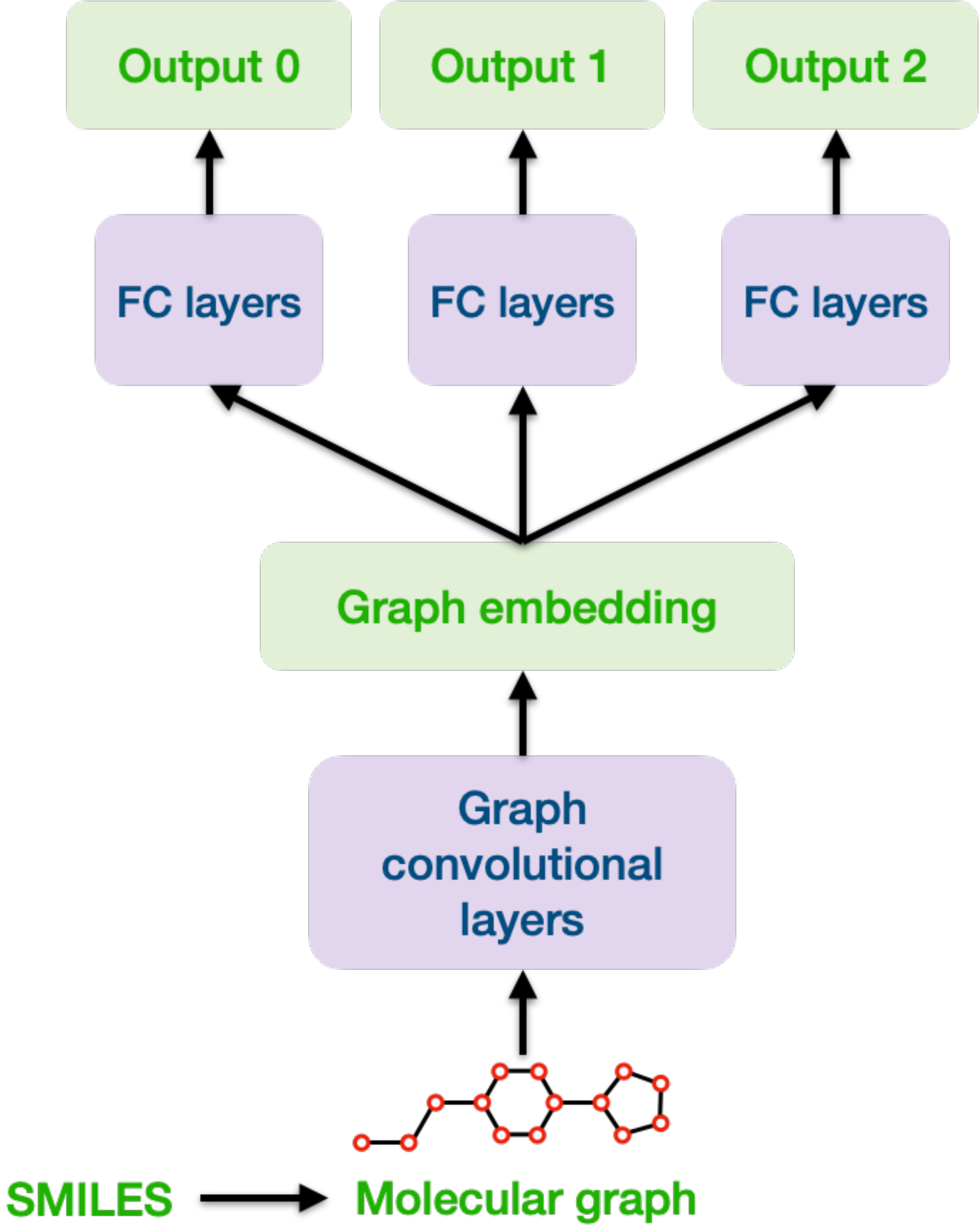}
\caption{The MTL model for virtual screening with graph neural networks. The weights of the graph convolutional layers are shared across all the tasks. For each task, the output embedding of the graph convolutional layers goes into task-specific fully connected layers to get the final prediction.}
\label{fig:mtl2}
\end{figure}

\subsubsection{Input molecular features}

In a chemical library, the compounds are usually represented by SMILES (Simplified Molecular Input Line Entry System)~\cite{weininger1988smiles}. 
Recently, modeling the molecules as graphs has shown a great success in many problems related to drug discovery, including molecular property prediction, drug-target interaction prediction, and molecule design~\cite{gilmer2017neural,liu2019chemi, sun2020graph, nguyen2021graphdta}. Thus we convert the SMILES codes to molecular graphs with the package RDKit~\cite{landrum2006rdkit}, such that a compound is a graph with the nodes being the atoms and the edges being the chemical bonds. RDKit is also used to generate features for the molecular graph, including seven atom features and two bond features. All the features are discrete values and encoded as one-hot vectors with different lengths, as shown in Table~\ref{tab1}. 


\begin{table}[htbp]
\caption{Input features for the molecular graph}
\begin{center}
\begin{tabular}{|c|l l|}
\hline
 & atomic type &119 bits one hot \\
 & formal charge& 16 bits one hot \\
Atom  & degree& 11 bits one hot \\
feature & chirality tag& 4 bits one hot \\
 & number of hydrogens& 9 bits one hot \\
 & aromaticity& 2 bits one hot \\
 & hybridization& 5 bits one hot  \\
\hline
Bond & bond direction,&7 bits one hot \\
feature & bond type& 4 bits one hot \\
 & is in ring& 2 bits one hot \\
\hline
\end{tabular}
\label{tab1}
\end{center}
\end{table}

\subsubsection{Parameter sharing module}
With the input of molecular graph data, the parameter-sharing part of our MTL model $G_{\phi}$ is based on the Graph Isomorphism Network (GIN), which has been shown to achieve the state-of-the-art performance in drug-target affinity prediction~\cite{xu2018how,nguyen2021graphdta}. 
As shown in Table~\ref{tab1}, the features generated by RDKit are all one-hot vectors, which are denoted as $f_{v,i}$ ($i=1,\ldots,7$) and $f_{e,j}$ ($j=1,2,3$) for node $v$ and edge $e$. In order to input the features to GNNs, we first use embedding operations to map the one-hot vectors into $d$-dimensional real vectors: 
\begin{align*}
    h_v^{0} & = \sum_{i=1}^7 \mathrm{NodeEmbedding}_i(f_{v,i}), \\
    h_e^{k} & = \sum_{j=1}^3  \mathrm{EdgeEmbedding}^{k}_{j}(f_{e,j}) \quad k=0,1,\ldots,K-1,
\end{align*}
where $K$ is the total number of GNN layers. In the $k$-th layer, GIN updates the node representations by
\begin{align*}
    h_v^{k} = \sigma
    \left( g^k \left ( \sum_{u\in \mathcal{N}(v) \cup \{ v \} } h_u^{k-1} + \sum_{e=(v,u): u\in \mathcal{N}(v) \cup \{ v \} } h_e^{k-1}
                \right )
    \right ),
\end{align*}
where $\mathcal{N}(v)$ is the set of nodes adjacent to node $v$, $\sigma(\cdot)$ is the ReLU activation function, and $g^k(\cdot)$ is a two-layer perceptron with $2d$ hidden neurons followed by batch normalization~\cite{ioffe2015batch}.
Since the docking score prediction is a graph-level regression task, the graph representation is obtained by averaging the node embeddings of the last GIN layer:
\begin{align*}
    z = \mathrm{MEAN}_{v}(h_v^{K}).
\end{align*}

\subsubsection{Task-specific module}
The output of the GIN layers gives a vector representation of the compounds, which contains shared knowledge across all the tasks. These vector representations are then fed into the task-specific neural networks to make predictions for each task. 
Here, we apply a two-layer fully connected (FC) neural network on top of the graph representation $z$ to obtain the final prediction for each task.

\subsection{Other methods for comparison}
The first baseline for comparison is the traditional single-task machine learning method for virtual screening.
In addition, several recent works for docking-based virtual screening have focused on the technique of active learning~\cite{graff2021accelerating,yang2021efficient}. We therefore also compare MTL with the (single-task) active learning approach.
For both cases, the same neural network architecture is used. The only difference is that the number of tasks is set to be one.
For active learning, we adopt the same data acquisition strategy as described in~\cite{yang2021efficient}, where an ensemble of five models with the same architecture is used. To ensure a fair comparison, the training samples of the new target task always have the same size for all the approaches.

\section{Results}
The performances of different machine learning approaches for virtual screening are evaluated with two large docking datasets.
In addition, we also collect a drug-target affinity dataset to illustrate that MTL is applicable to other problems for drug discovery.

\subsection{Experiment details}
\subsubsection{Dataset}
The first docking dataset was downloaded from ~\cite{gentile2020deep}, which contained 12 targets (PDB id: 1ERR, 1T7R, 2ZV2, 4AG8, 4F8H, 4R06, 4YAY, 5EK0, 5L2S, 5MZJ, 6D6T, 6IIU), and each target was docked with 3 million compounds randomly sampled from ZINC15 by the program FRED~\cite{mcgann2012fred}.
The second dataset was obtained from~\cite{lyu2019ultra}, in which 99 million lead-like compounds were docked against the target AmpC (AmpC $\beta$-lactamase) using DOCK3.7.2~\cite{coleman2013ligand}. 
These two datasets allow us to test the performance of MTL for virtual screening in a large scale and across different docking softwares.
To further test the ability of MTL in drug discovery, we also collect an experimental drug-target affinity dataset (the half-maximal inhibitory concentrations of the ligands on their protein targets, i.e., $\mathrm{IC}_{50}$) with eight targets from the ChEMBL database~\cite{mendez2019chembl}. ChEMBL uses the pChEMBL value as the measurement of the ligand-target affinity, which is defined as:
\begin{equation}
    \mathrm{pChEMBL} = -\log (\mathrm{IC}_{50}).
\end{equation}
For example, an $\mathrm{IC}_{50}$ measurement of 10 uM would have a pChEMBL value of 5. We use this pChEMBL value as the ground truth in our experiment. When a ligand-target pair has multiple pChEMBL values, we take the average of all the pChEMBL values. The details of the drug-target affinity dataset are summarized in TABLE~\ref{tab_chembl}.

\begin{table}[htbp]
\caption{Details of the drug-target affinity dataset collected from ChEMBL}
\begin{center}
\begin{tabular}{|c|l | c|}
\hline
ChEMBL ID & Target name & Data points \\
\hline
\hline
203 & Epidermal growth factor receptor erbB1 & 6414 \\
\hline
279 & Fetal liver kinase 1 & 7777 \\
\hline
267 & Tyrosine-protein kinase SRC & 2775 \\
\hline
325 & Histone deacetylase 1 & 4473 \\
\hline
333 & Matrix metalloproteinase-2 & 2673 \\
\hline
2842 & Serine/threonine-protein kinase mTOR & 3662 \\
\hline
2971 & Tyrosine-protein kinase JAK2 & 4855 \\
\hline
4005 & PI3-kinase p110-alpha subunit & 4459 \\
\hline
\end{tabular}
\label{tab_chembl}
\end{center}
\end{table}

\begin{figure*}[ht!]
\centering
     \begin{subfigure}[b]{0.235\textwidth}
         \centering
         \includegraphics[width=\textwidth]{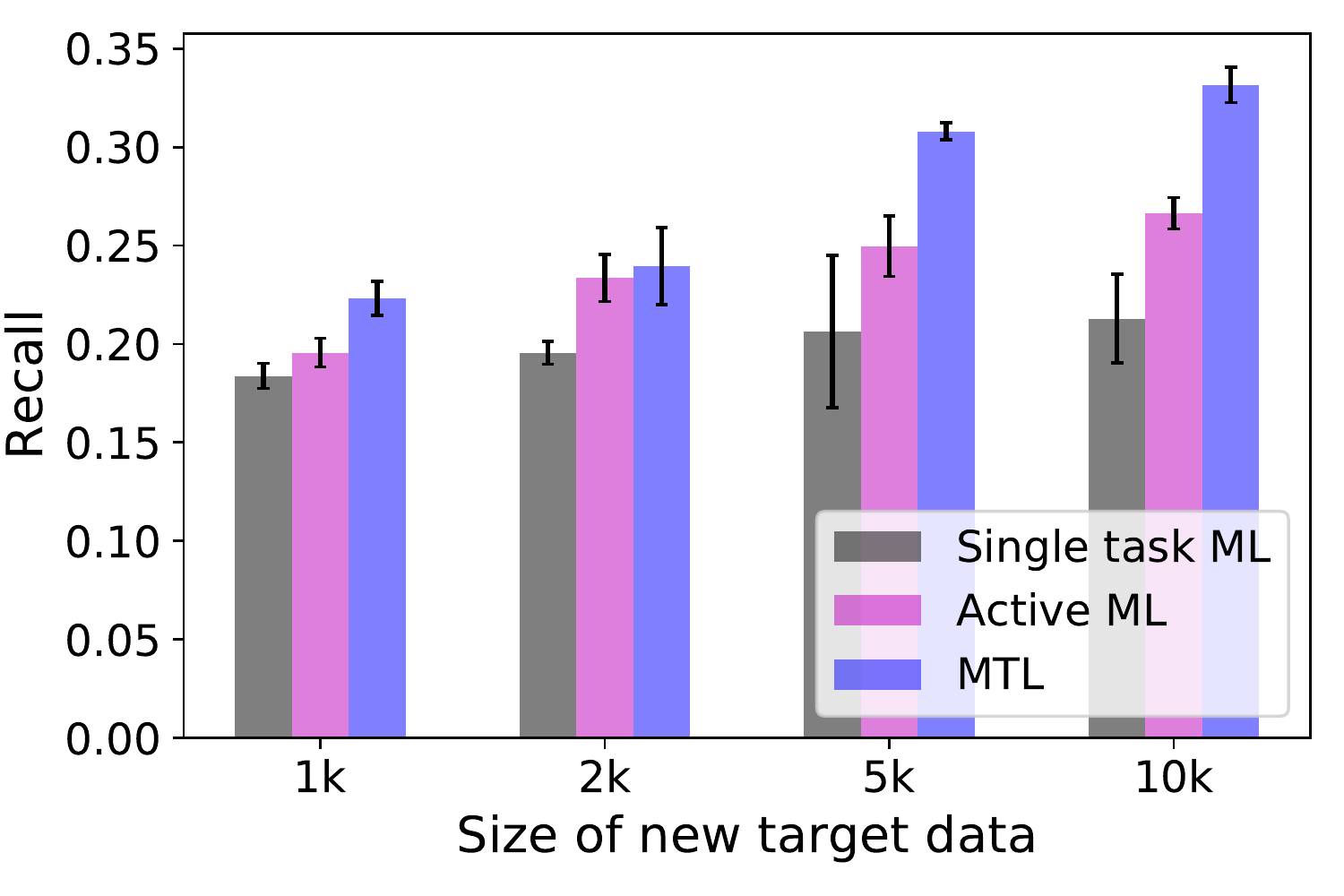}
         \caption{1ERR}
         \label{fig:MTL1}
     \end{subfigure}~
     ~
     \begin{subfigure}[b]{0.235\textwidth}
         \centering
         \includegraphics[width=\textwidth]{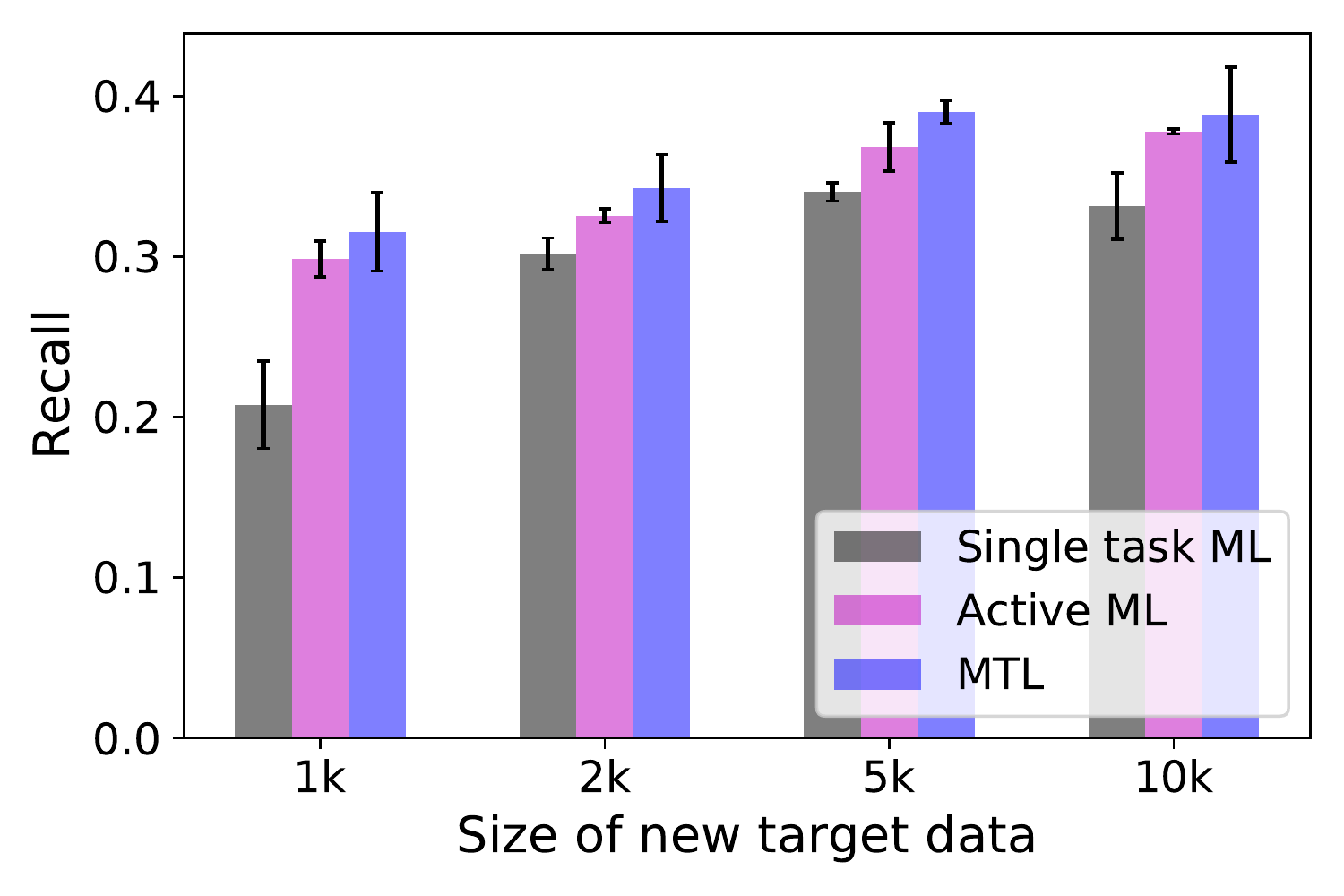}
         \caption{1T7R}
         \label{fig:MTL2}
    \end{subfigure}
    ~
    \begin{subfigure}[b]{0.235\textwidth}
         \centering
         \includegraphics[width=\textwidth]{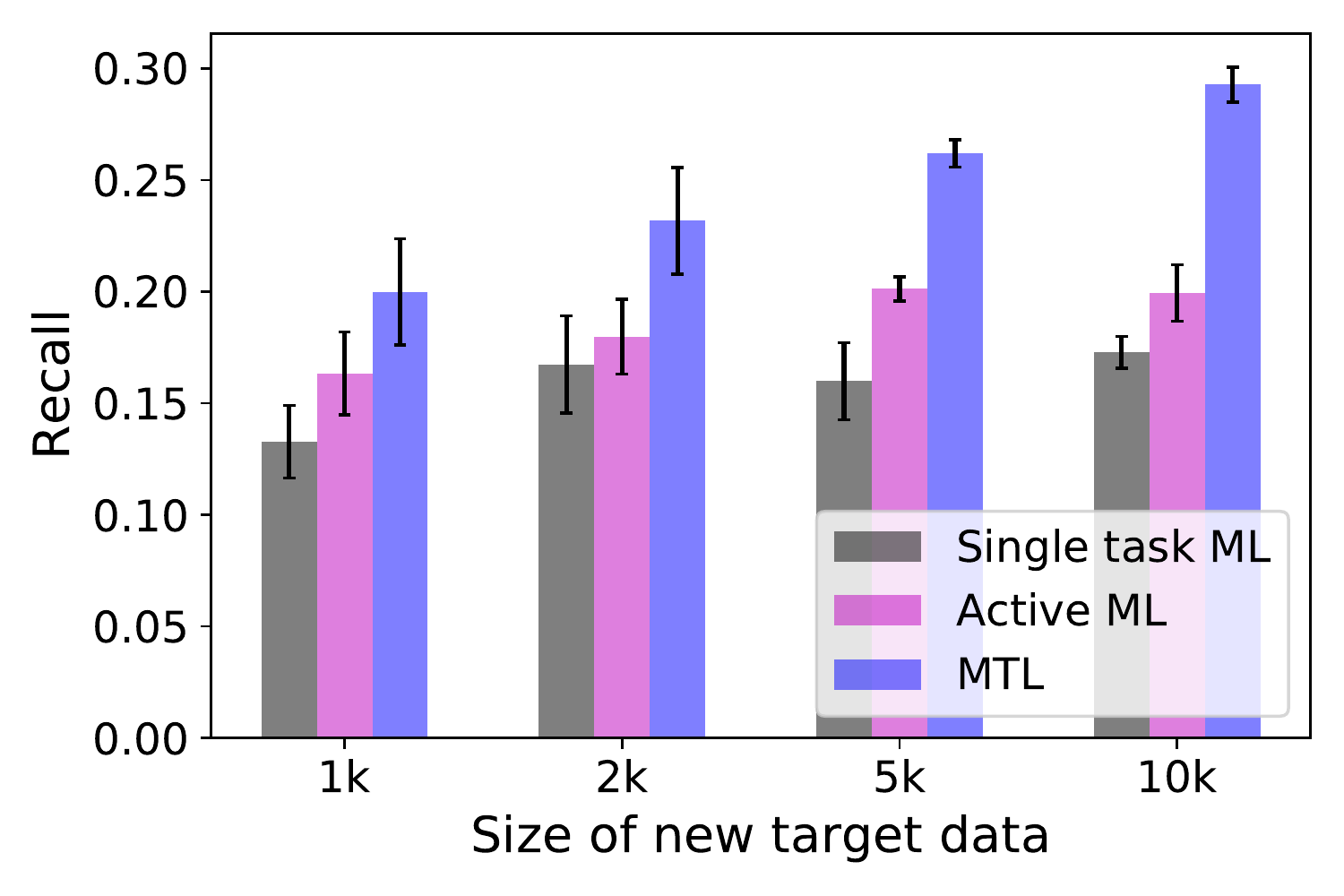}
         \caption{2ZV2}
         \label{fig:MTL3}
     \end{subfigure}
     ~
    \begin{subfigure}[b]{0.235\textwidth}
         \centering
         \includegraphics[width=\textwidth]{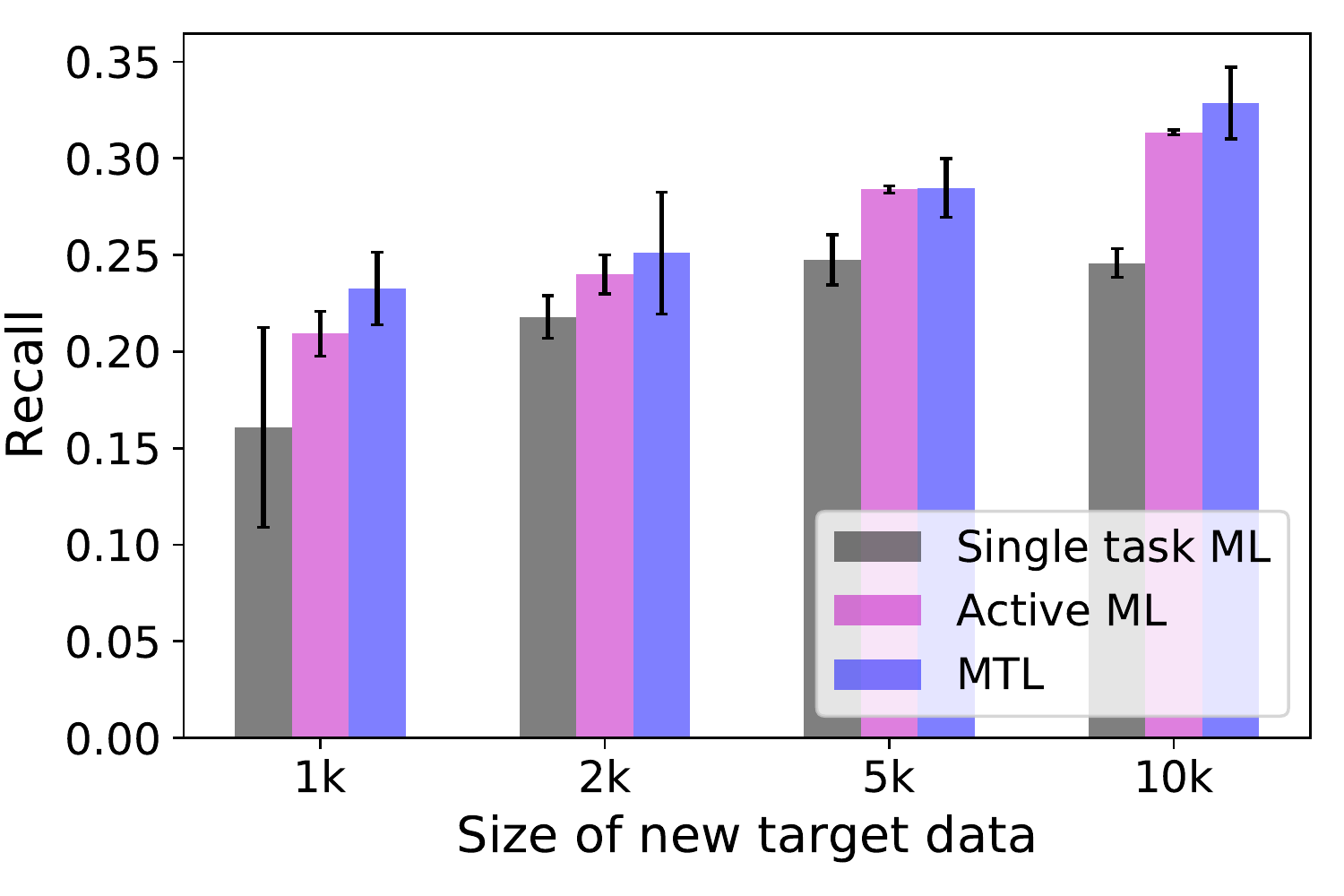}
         \caption{4AG8}
         \label{fig:MTL4}
     \end{subfigure}
     ~
    \begin{subfigure}[b]{0.235\textwidth}
         \centering
         \includegraphics[width=\textwidth]{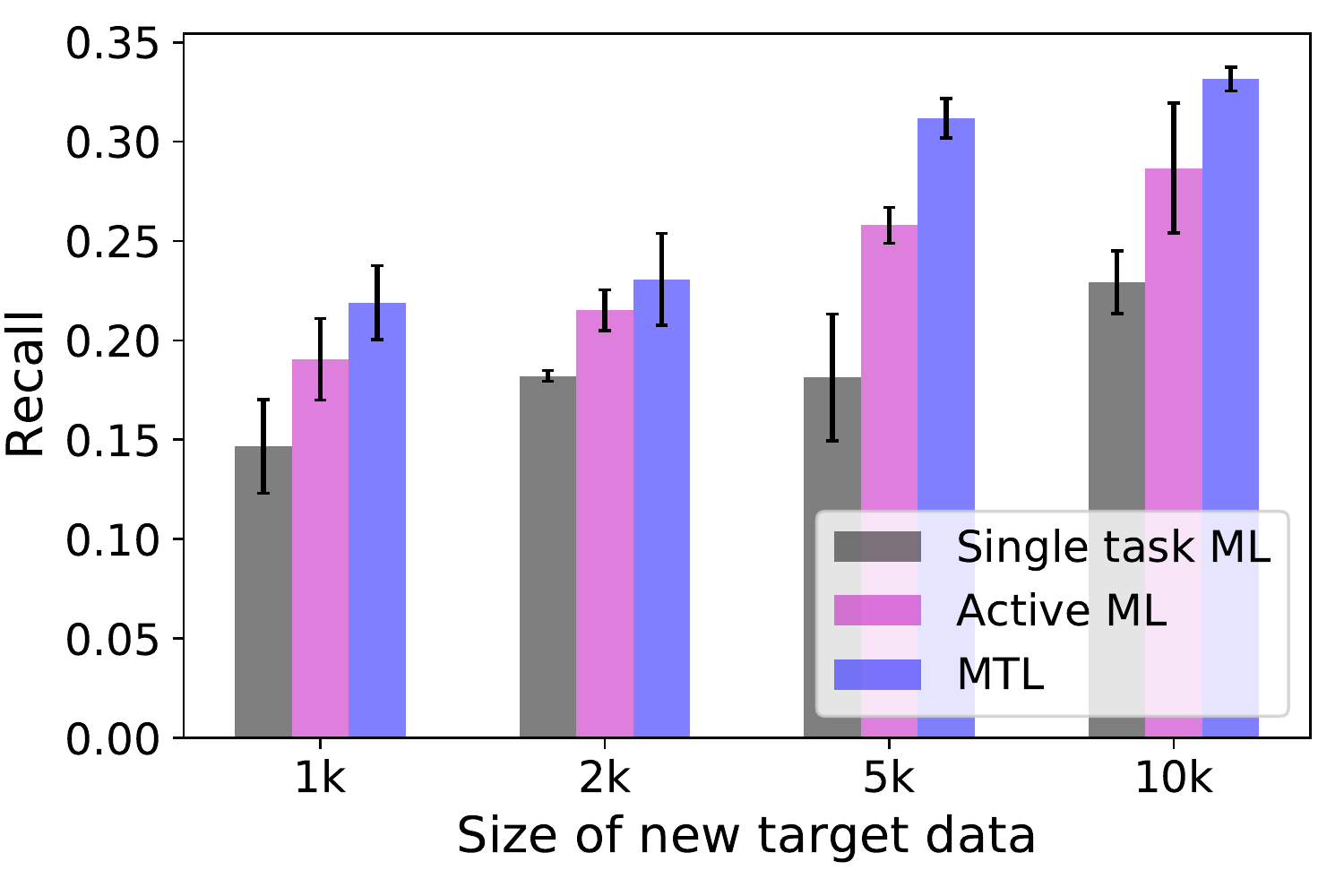}
         \caption{4F8H}
         \label{fig:MTL5}
     \end{subfigure}
     ~
    \begin{subfigure}[b]{0.235\textwidth}
         \centering
         \includegraphics[width=\textwidth]{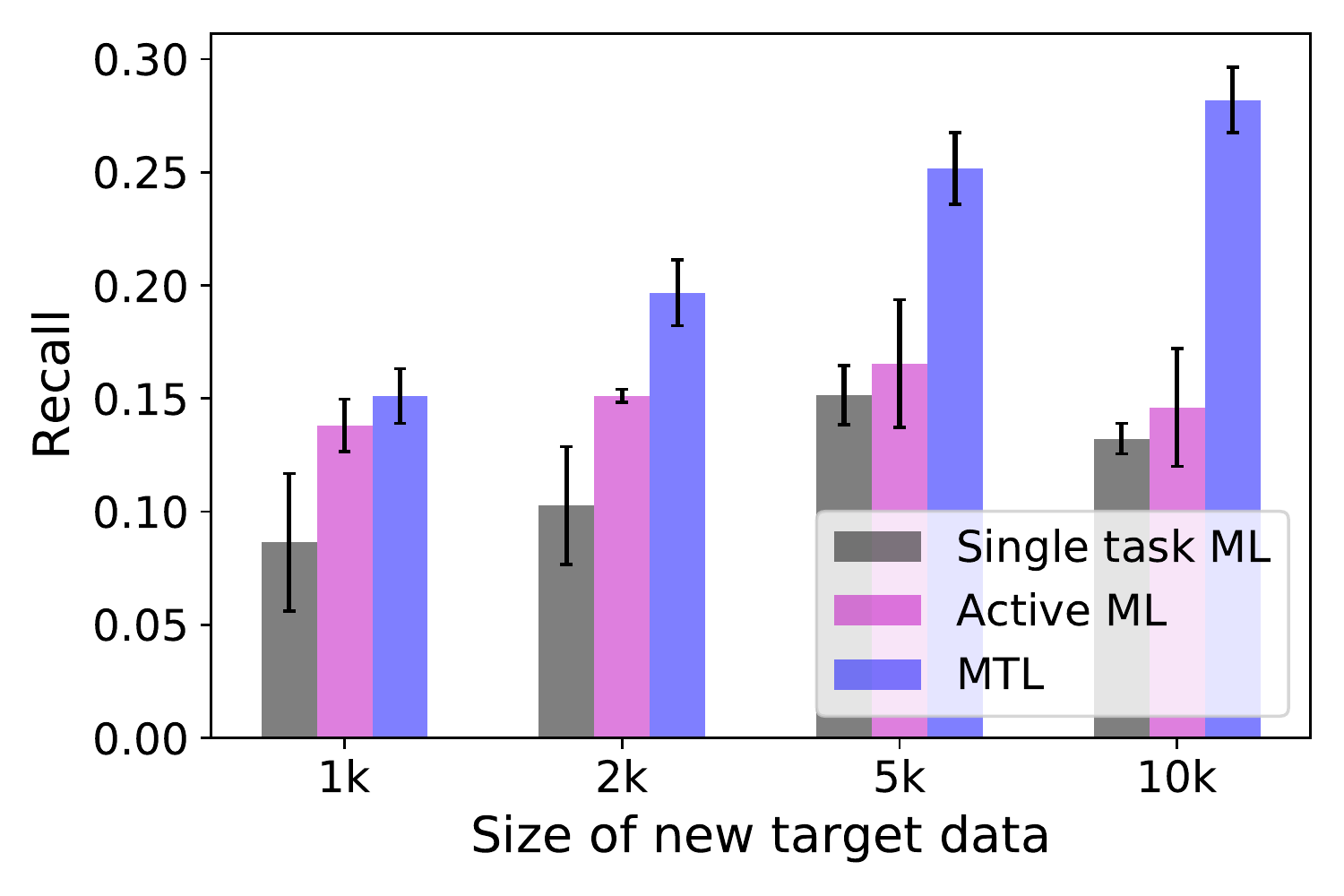}
         \caption{4R06}
         \label{fig:MTL6}
     \end{subfigure}
     ~
    \begin{subfigure}[b]{0.235\textwidth}
         \centering
         \includegraphics[width=\textwidth]{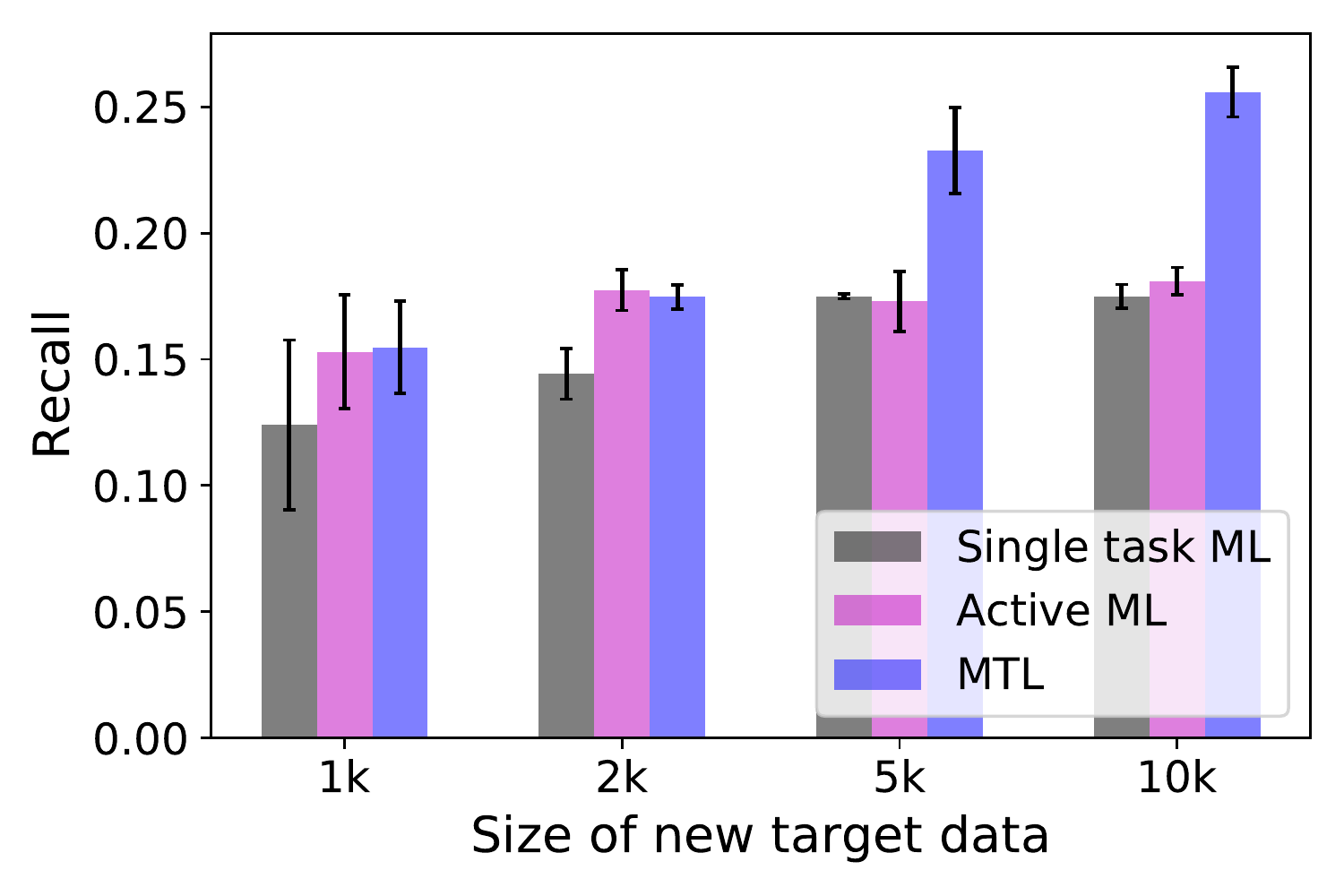}
         \caption{4YAY}
         \label{fig:MTL7}
     \end{subfigure}
     ~
    \begin{subfigure}[b]{0.235\textwidth}
         \centering
         \includegraphics[width=\textwidth]{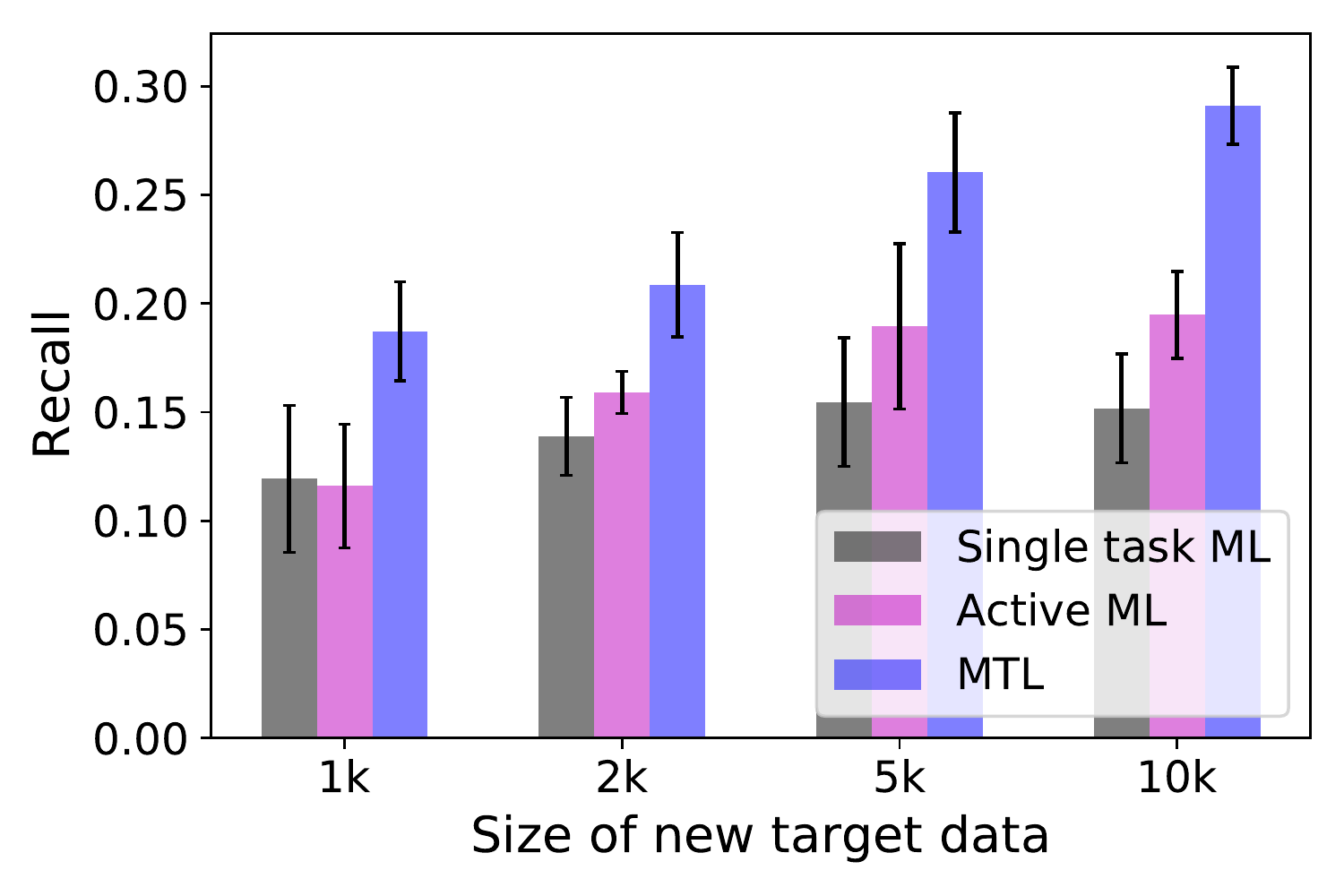}
         \caption{5EK0}
         \label{fig:MTL8}
     \end{subfigure}
     ~
    \begin{subfigure}[b]{0.235\textwidth}
         \centering
         \includegraphics[width=\textwidth]{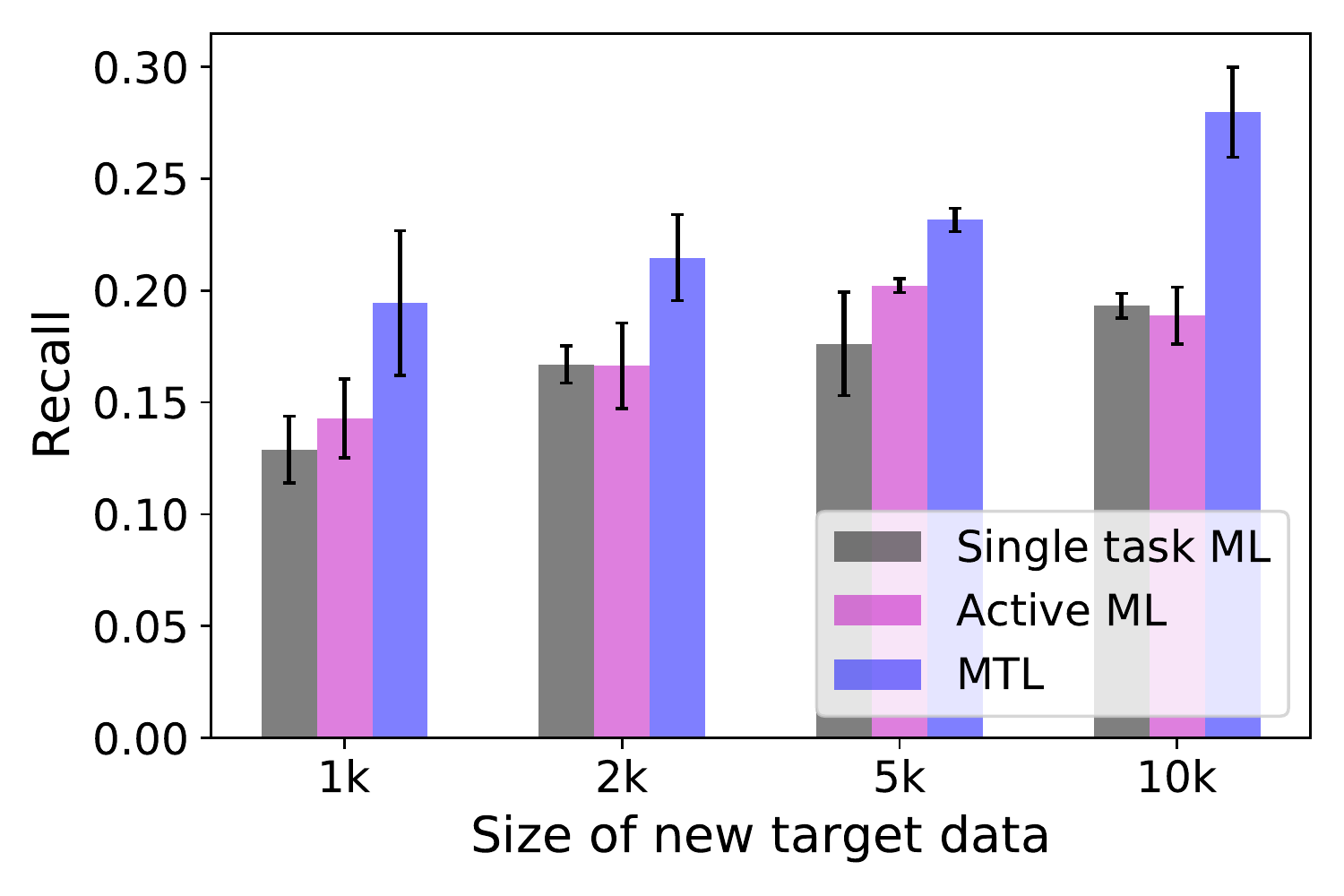}
         \caption{5L2S}
         \label{fig:MTL9}
     \end{subfigure}
     ~
    \begin{subfigure}[b]{0.235\textwidth}
         \centering
         \includegraphics[width=\textwidth]{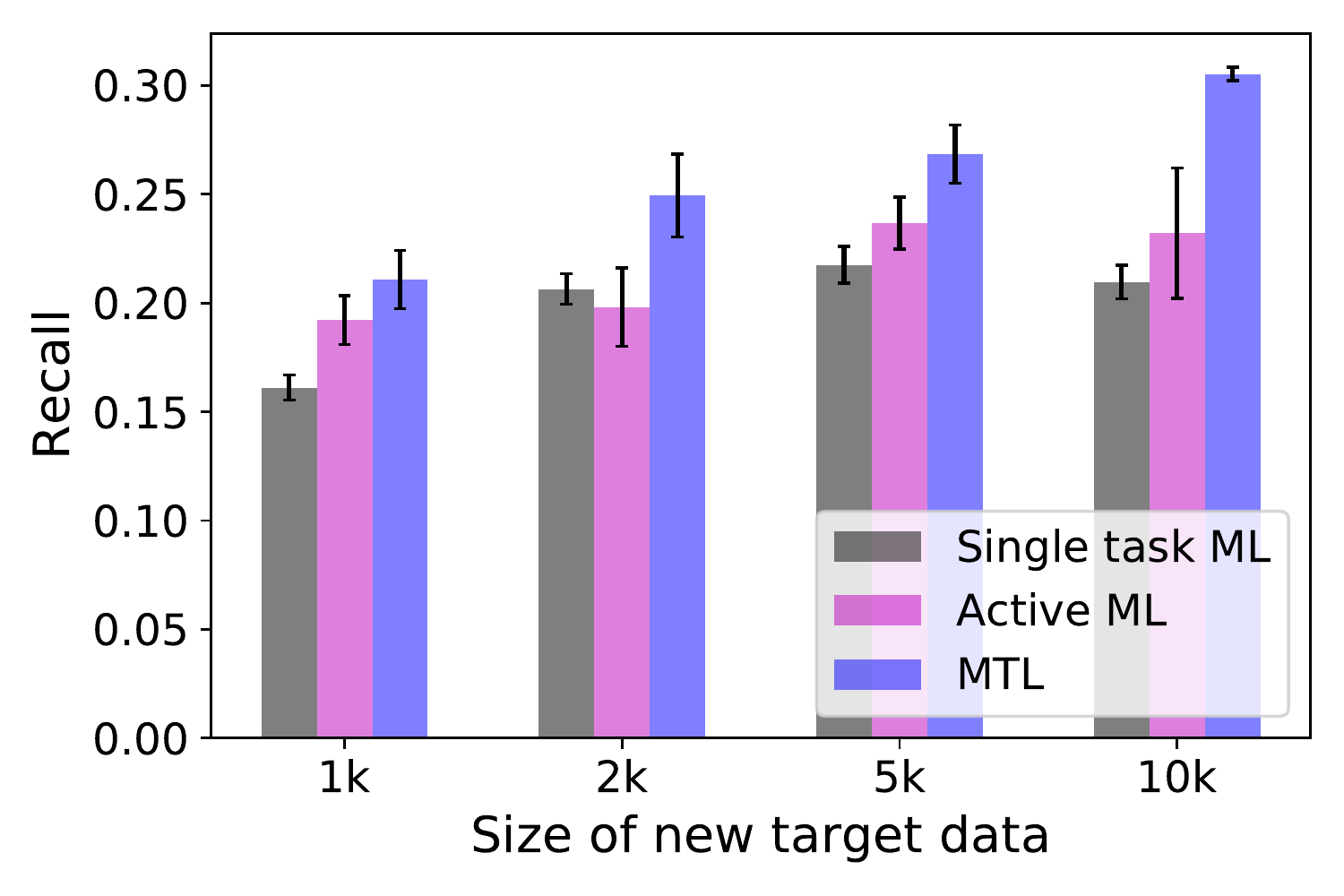}
         \caption{5MZJ}
         \label{fig:MTL10}
     \end{subfigure}
     ~
    \begin{subfigure}[b]{0.235\textwidth}
         \centering
         \includegraphics[width=\textwidth]{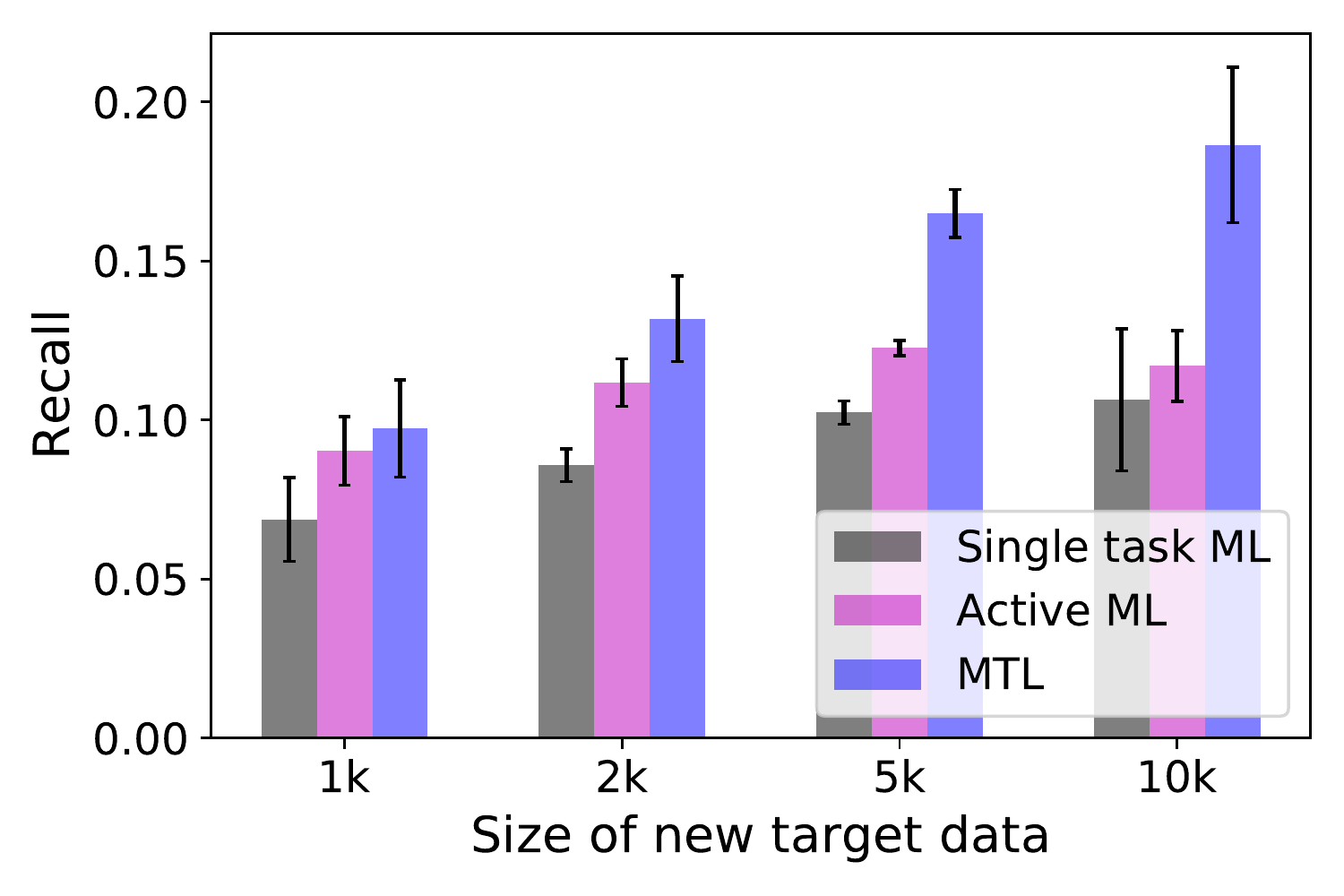}
         \caption{6D6T}
         \label{fig:MTL11}
     \end{subfigure}
     ~
    \begin{subfigure}[b]{0.235\textwidth}
         \centering
         \includegraphics[width=\textwidth]{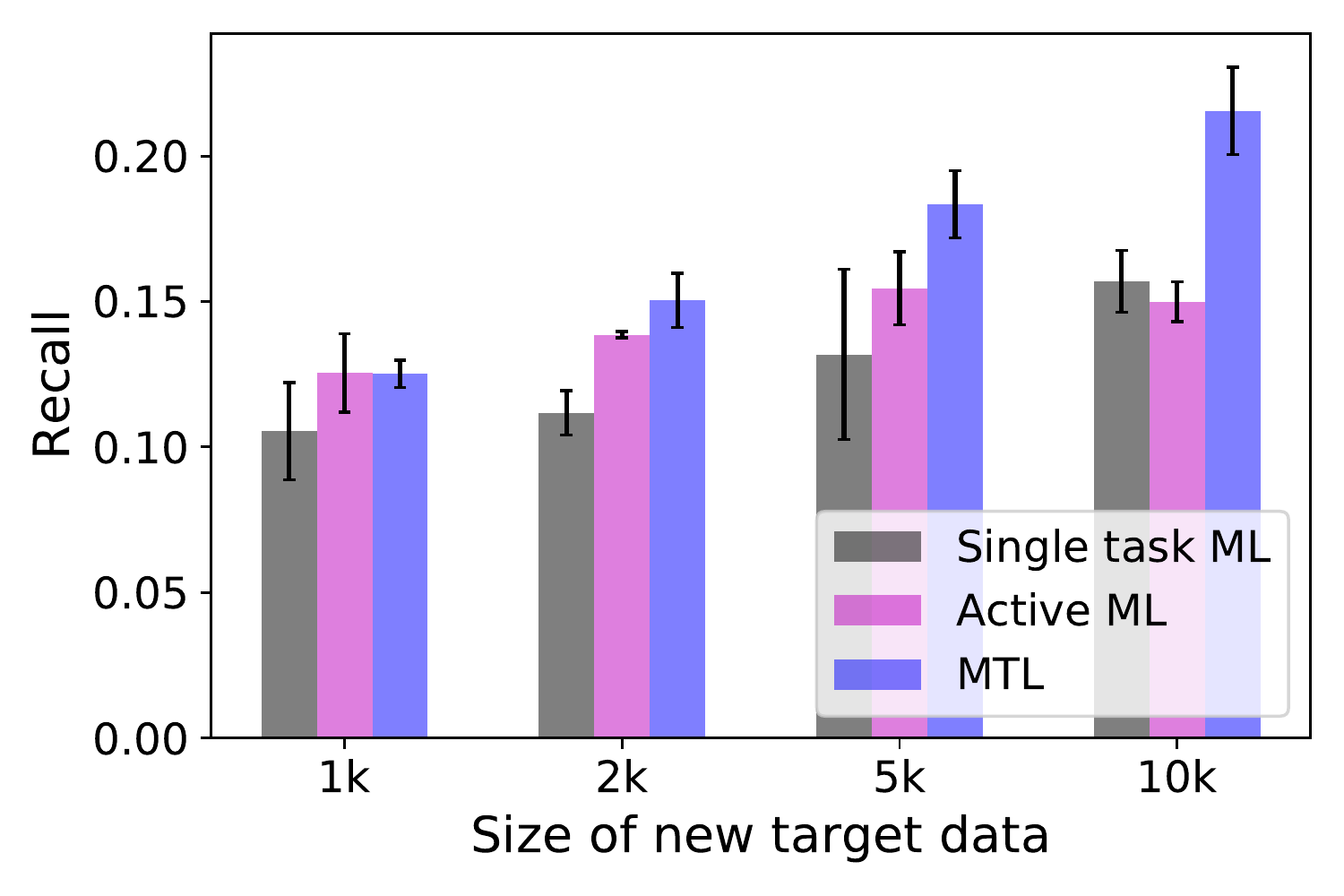}
         \caption{6IIU}
         \label{fig:MTL12}
     \end{subfigure}
\caption{Recall of top 3k virtual hit of three approaches with different `new target' data sizes for all 12 targets at top $2\%$ predicted values. For MTL, the `new target' data is trained together with 2 million samples from the 11 `developed targets'. The error bars are obtained from three independent runs.}
\label{fig:MTL}
\end{figure*}

\subsubsection{Evaluation metrics}
The performance of the machine learning approaches for docking score prediction is evaluated by the virtual hit ratio. 
We define the top-$k$ compounds as the virtual hits. For the first dataset of 3 million compounds, $k$ is set to be 3,000 or 30,000. For the larger dataset of 99 million compounds, $k$ is 10,000 or 100,000.
The predicted virtual hits are taken as the top $2\%$, $3\%$ or $5\%$ compounds of the predictions.
The hit ratio is calculated as the recall:
\begin{align*}
    \mathrm{recall} = \frac{TP}{TP+FN},
\end{align*}
where TP (true positives) are the correctly predicted virtual hits, and FN (false negatives) are the virtual hits discarded by the prediction incorrectly.

The performance of the drug-target affinity prediction is measured by the mean-square error (MSE) and concordance index (CI). The CI is defined as
\begin{align*}
    \mathrm{CI} = \frac{\sum_{k,l} \mathbf{1}_{\Tilde{y}_k > \Tilde{y}_l}\cdot \mathbf{1}_{y_k > y_l}}
    {\sum_{k,l} \mathbf{1}_{y_k > y_l}},
\end{align*}
where $\Tilde{y}_k$ and $y_k$ are the predicted and true value of the $k$-th sample respectively, and $\mathbf{1}_{{y}_k > {y}_l}=1$ if ${y}_k > {y}_l$, else $0$.

\subsubsection{Model training details}
The embedding dimension for the features $d$ is $256$. The number of GIN layers is eight. The number of hidden neurons in the task-specific FC layers is also $256$. The model is trained with the ADAM optimizer with a learning rate $0.001$ and batch size $128$~\cite{kingma2014adam}. During training, we apply dropout after the ReLU for all GNN layers and the task-specific FC layers~\cite{srivastava2014dropout}. The dropout rate is set to be $0.2$. $20\%$ of the training data are used for validation to avoid over-fitting. The model is trained for at least $100$ epochs until the validation loss is larger than the training loss for more than $50$ consecutive epochs. 
The epoch with the smallest validation loss is chosen as the final model.
For active learning, the final prediction is the average of the five models in the ensemble.

\subsection{MTL for docking-based virtual screening}
For the first dataset, one of the 12 targets is taken as the `new target' and the other 11 targets are treated as the `developed targets' with existing data. For each new target, we randomly choose 1k, 2k, 5k or 10k samples as the training data. For active learning, the total number of samples for training is kept to be the same to ensure a fair comparison. 
For MTL, we choose 100k, 500k, 1m (million), and 2m samples randomly from the 11 developed targets to jointly train the model with the samples from the new target. \autoref{fig:MTL} shows the bar plots comparing the recalls of the top 3000 virtual hits of the top $2\%$ of the compounds predicted by three machine learning approaches across different new target data sizes.
As all three approaches have better performances when the training size of the new target samples increases, MTL achieves higher recall values than single-task and active learning in all the targets across four different new target data sizes. In several targets (such as 2ZV2 and 5EK0), MTL with 1k new target data can obtain a comparable recall value to the single-task machine learning and active learning with 10k new target data, which means that MTL requires less docking data of the new target to achieve the same performance. 
Thus in practice, MTL can further accelerate docking-based virtual screening without sacrificing the performance.
As a summary of the prediction performance, TABLE~\ref{tab_pearson} gives the Pearson correlation between the real and predicted docking scores.
The complete results of all the targets across different new target sizes and other target sizes for MTL can be found in the Supplementary Table~\footnote{\url{https://github.com/PaddlePaddle/PaddleHelix/tree/dev/apps/drug_target_interaction/MTL_docking/BIBM_SupplementaryMaterials}}. 

\begin{figure}[!ht]
\centering
\includegraphics[width=0.7\linewidth]{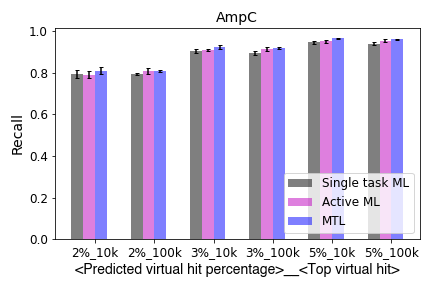}
\caption{Recall of the top 10k and 100k virtual hits of different percentages of the predicted values of the 99 million AmpC docking data. Although single-task and active learning already achieve very high recall values, MTL performs better.}
\label{fig:ampc}
\end{figure}

To further examine the ability of MTL for large-scale virtual screening, we use the second dataset, which contains the docking scores of 99 million compounds against the target AmpC.
Out of the 99 million docking results, 100k samples of AmpC are chosen as the new target data. 2m random samples from the 12 targets in the first dataset are used in MTL as the data of other tasks.
As shown in~\autoref{fig:ampc}, MTL performs the best when evaluated across different numbers of virtual hits or predicted virtual hits.
Note that the AmpC docking data is produced by a different docking software. So our result also indicates that MTL allows knowledge-sharing across different docking programs.

\begin{table*}[ht!]
\caption{Pearson correlation of single-task machine learning (ML), active learning (AL) and MTL with different new target sizes. MTL is jointly trained with 2 million data from other targets.}
\centering
\begin{tabular}{l|lll|lll|lll|lll|}
\cline{2-13}
                           & \multicolumn{3}{c|}{Target size=1k} & \multicolumn{3}{c|}{Target size=2k}      & \multicolumn{3}{c|}{Target size=5k}      & \multicolumn{3}{c|}{Target size=10k}     \\ \cline{2-13} 
                           & ML      & AL      & MTL             & ML    & AL    & \multicolumn{1}{c|}{MTL} & ML    & AL    & \multicolumn{1}{l|}{MTL} & ML    & AL    & \multicolumn{1}{l|}{MTL} \\ \hline
\multicolumn{1}{|l|}{1ERR} & 0.562   & 0.566   & \textbf{0.630}  & 0.605 & 0.610 & \textbf{0.662}           & 0.617 & 0.635 & \textbf{0.710}           & 0.627 & 0.645 & \textbf{0.732}           \\
\multicolumn{1}{|l|}{1T7R} & 0.800   & 0.790   & \textbf{0.844}  & 0.823 & 0.826 & \textbf{0.859}           & 0.851 & 0.852 & \textbf{0.882}           & 0.859 & 0.854 & \textbf{0.879}           \\
\multicolumn{1}{|l|}{2ZV2} & 0.531   & 0.545   & \textbf{0.618}  & 0.581 & 0.582 & \textbf{0.646}           & 0.588 & 0.606 & \textbf{0.680}           & 0.608 & 0.610 & \textbf{0.696}           \\
\multicolumn{1}{|l|}{4AG8} & 0.671   & 0.685   & \textbf{0.751}  & 0.718 & 0.737 & \textbf{0.771}           & 0.758 & 0.769 & \textbf{0.790}           & 0.770 & 0.777 & \textbf{0.812}           \\
\multicolumn{1}{|l|}{4F8H} & 0.566   & 0.596   & \textbf{0.658}  & 0.636 & 0.628 & \textbf{0.689}           & 0.650 & 0.668 & \textbf{0.721}           & 0.667 & 0.694 & \textbf{0.747}           \\
\multicolumn{1}{|l|}{4R06} & 0.493   & 0.543   & \textbf{0.597}  & 0.548 & 0.585 & \textbf{0.639}           & 0.610 & 0.602 & \textbf{0.680}           & 0.603 & 0.596 & \textbf{0.697}           \\
\multicolumn{1}{|l|}{4YAY} & 0.528   & 0.553   & \textbf{0.550}  & 0.568 & 0.580 & \textbf{0.589}           & 0.585 & 0.590 & \textbf{0.639}           & 0.595 & 0.601 & \textbf{0.669}           \\
\multicolumn{1}{|l|}{5EK0} & 0.492   & 0.496   & \textbf{0.564}  & 0.525 & 0.532 & \textbf{0.587}           & 0.564 & 0.560 & \textbf{0.650}           & 0.570 & 0.580 & \textbf{0.680}           \\
\multicolumn{1}{|l|}{5L2S} & 0.474   & 0.496   & \textbf{0.579}  & 0.519 & 0.516 & \textbf{0.620}           & 0.562 & 0.563 & \textbf{0.655}           & 0.586 & 0.527 & \textbf{0.679}           \\
\multicolumn{1}{|l|}{5MZJ} & 0.663   & 0.677   & \textbf{0.743}  & 0.704 & 0.704 & \textbf{0.762}           & 0.725 & 0.744 & \textbf{0.778}           & 0.733 & 0.744 & \textbf{0.798}           \\
\multicolumn{1}{|l|}{6D6T} & 0.498   & 0.534   & \textbf{0.616}  & 0.587 & 0.590 & \textbf{0.655}           & 0.611 & 0.620 & \textbf{0.694}           & 0.610 & 0.623 & \textbf{0.715}           \\
\multicolumn{1}{|l|}{6IIU} & 0.585   & 0.609   & \textbf{0.670}  & 0.639 & 0.648 & \textbf{0.708}           & 0.669 & 0.679 & \textbf{0.747}           & 0.694 & 0.667 & \textbf{0.764}          \\ \hline
\end{tabular}
\label{tab_pearson}
\end{table*}

\begin{table*}[h]
\caption{The MSE and CI of the drug-target affinity prediction. The standard deviation in the parentheses is computed from three independent runs.}
\begin{tabular}{lllllllll}
\multicolumn{9}{c}{MSE} \\ \hline
ChEMBL ID      & 325            & 333            & 4005           & 2971           & 2842           & 267            & 203            & 279            \\ \hline
Single-task ML & 0.779(0.017) & 1.173(0.017) & 0.632(0.047) & 0.739(0.004) & 0.624(0.010)  & 0.907(0.017) & 1.033(0.036) & 0.875(0.026) \\
MTL            & 0.646(0.019) & 0.960(0.035) & 0.480(0.022) & 0.586(0.017) & 0.477(0.006) & 0.775(0.010)   & 0.693(0.004) & 0.606(0.014) \\
Improvement           & 17.09\%        & 18.17\%        & 24.12\%        & 20.73\%        & 23.64\%        & 14.60\%        & 32.92\%        & 30.78\%        \\ \hline
    &                &                &                &                &               &                &                &                \\
\multicolumn{9}{c}{CI} \\ \hline
ChEMBL ID      & 325            & 333            & 4005           & 2971           & 2842           & 267            & 203            & 279            \\ \hline
Single-task ML & 0.721(0.011) & 0.755(0.005) & 0.756(0.007) & 0.720(0.011) & 0.775(0.005) & 0.761(0.003) & 0.729(0.005) & 0.691(0.012)  \\ 
MTL            & 0.753(0.005) & 0.785(0.005) & 0.793(0.003) & 0.765(0.004) & 0.810(0.001) & 0.779(0.003) & 0.789(0.001) & 0.762(0.004) \\
Improvement      & 4.41\%         & 3.90\%         & 4.97\%         & 6.33\%         & 4.50\%         & 2.31\%         & 8.19\%         & 10.16\%    \\ \hline   
\end{tabular}
\label{tab_dta}
\end{table*}

\subsection{Results on drug-target affinity prediction}
Next, we use the collected drug-target affinity dataset to further test the ability of MTL in other problems of drug discovery. As in the first docking dataset, we take one of the targets as the new target, of which $60\%$ samples are used as the testing set for evaluation. The other $40\%$ samples are used for single-task machine learning, or combined with the data of the other targets for MTL training. TABLE~\ref{tab_dta} shows that compared to single-task machine learning, MTL has a substantial improvement in terms of both MSE and CI, indicating that MTL helps predict experimental drug-target affinity values and is applicable to not only docking score prediction but also other tasks in drug discovery.

To obtain an intuitive understanding of the difference between the single and multi-task learning models, we visualize the compound embedding learned by the GIN layers. 
Specifically, we randomly sample 1000 compounds for each of the 10 most common scaffolds from the ZINC database~\cite{bemis1996properties}.
UMAP is then used to visualize the embedding of these 10000 compounds from the last GIN layer~\cite{mcinnes2020umap}. As shown in~\autoref{fig:umap}, the model trained by MTL gives more distinctive clusters with correspondence to the 10 scaffolds than the single-task model. Quantitatively, the embedding of the MTL model has a smaller DB index and higher Silhouette index~\cite{xie1991validity,rousseeuw1987silhouettes}.
As the scaffold represents the core structure of a compound, our results indicate that the model trained with MTL captures more inherent characteristics via sharing knowledge across multiple tasks. 
By MTL, the learned model achieves a similar effect of representation learning by self-supervised learning~\cite{li2021effective}.
Thus the high-quality representation learned by MTL leads to greater predictive power compared to single-task learning.

\begin{figure}[!ht]
\centering
\includegraphics[width=0.75\linewidth]{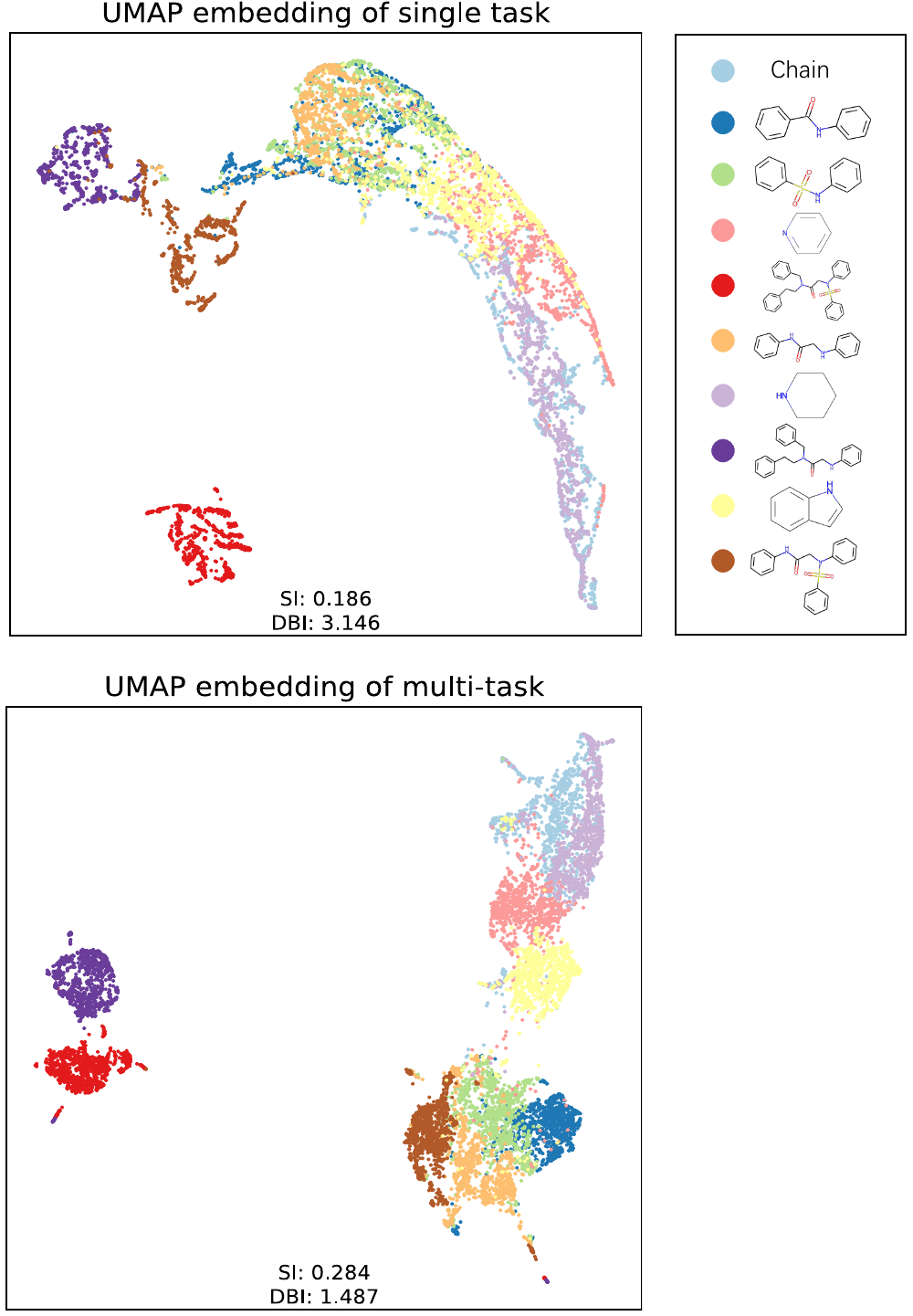}
\caption{Visualization of the compound embeddings extracted from the final layer of the GINs from the trained single-task and multi-task models. A lower DB index or higher Silhouette index indicates a more appropriate separation.}
\label{fig:umap}
\end{figure}

\begin{figure}[!ht]
\centering
\includegraphics[width=0.7\linewidth]{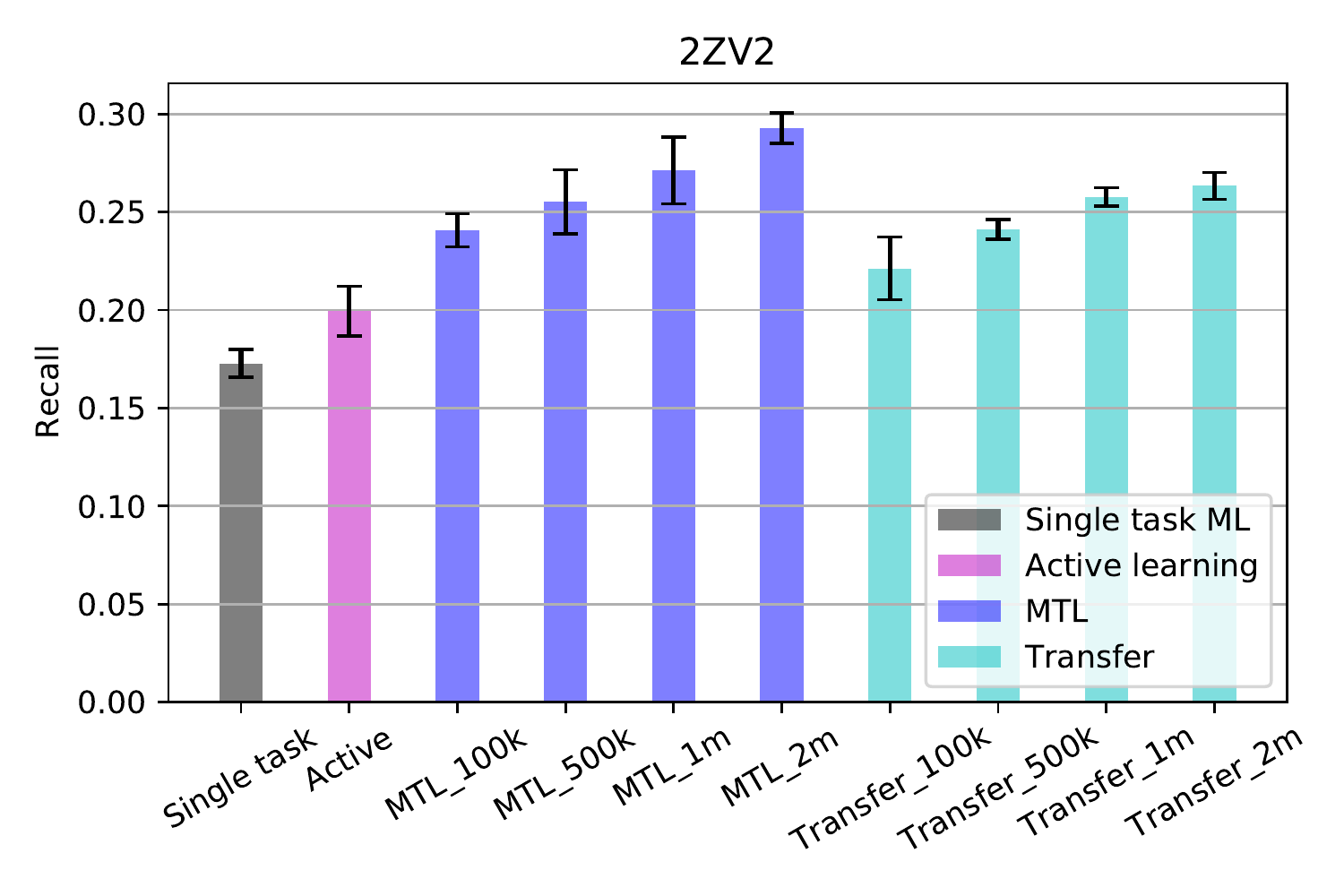}
\caption{Bar plots of the recall of top 3k virtual hit for the target 2ZVS at top $2\%$ predicted values of different ML approaches. The training size for the new target is 10k. MTL and transfer learning include the results of 4 different training sizes for the other targets. The error bars are obtained from three independent runs.}
\label{fig:transfer}
\end{figure}

\subsection{MTL trained model contains common reusable knowledge}
We have shown that MTL helps learn a good representation of the chemical compounds, which indicates that the model is able to capture common knowledge across all the tasks.
We next examine if the knowledge learned by MTL can be reused for a new task by transferring the MTL trained model to a new target and checking if it can improve the prediction performance. 
We test this idea using the first docking dataset with 12 targets. Similarly, one target is taken as the new target. An MTL model is first trained with the data of the other 11 targets. The weights of the MTL trained GIN layers are used as the initialization to train a single-task model of the new target. We first train the FC layers for 20 epochs and then train the whole neural network. From~\autoref{fig:transfer}, it is clear to see that transfer learning based on the MTL model achieves better recall values than single-task machine learning and active learning.~\autoref{fig:transfer} also reports that when the data size of the other tasks increases from 100k to 2m, the performances of MTL and MTL initialized transfer learning become better.

\section{Conclusion}
As the number of compounds in chemical libraries available to screening grows rapidly, machine learning approaches play an important role in docking-based virtual screening. Current works for docking barely consider using the docking data of other targets from previous screens. 
Our work is motivated by a practical question of how to utilize the existing data with machine learning during the development of a new drug target. 
With this motivation, we investigate MTL for the task of docking-based virtual screening in drug discovery. Combining the data from other tasks, we show that MTL achieves better performances than single-task and active machine learning by sharing knowledge across multiple tasks. Further experiments on a collected drug-target affinity dataset indicate that MTL learns better representations of the compounds and shows the potential in other problems for drug discovery. For example, it is straightforward to extend the MTL framework to applications of more computational costly jobs such as free energy calculation~\cite{wang2015accurate}.

In this work, we focus on the application of MTL in docking-based virtual screening. In the aspect of deep learning model development, a lot of efforts have been made to design new neural network architectures for MTL. One direction of future works can be optimization of the MTL neural network architecture for virtual screening in drug discovery.
Regarding how to make use of additional data, various other machine learning techniques can also be investigated in the future, such as meta-learning, self-supervised learning, and semi-supervised learning, which are all closely related to MTL~\cite{hospedales2021meta,van2020survey,liu2021self}.



\bibliographystyle{./IEEEtran}
\bibliography{./IEEEfull,./ref}

\end{document}